\documentclass{ifacconf}

\usepackage{graphicx}      
\usepackage{natbib}        
\usepackage{moreverb,url}
\usepackage{subcaption}
\usepackage[justification=centering]{caption}
\usepackage{mathtools}
\usepackage{amsmath,amsfonts,amssymb}
\usepackage{newpxtext} 
\usepackage[normalem]{ulem}

\begin{document}
\begin{frontmatter}

\title{Pendulum Actuated Spherical Robot: Dynamic Modeling \& Analysis for
Wobble \& Precession \thanksref{footnoteinfo}} 

\thanks[footnoteinfo]{This work is funded by a project under NCETIS (National
Center of Excellence in Technology and Internal Security),
IIT Bombay sponsored by Ministry of Electronics and
Information Technology.}

\author[First]{Animesh Singhal} 
\author[Second]{Sahil Modi} 
\author[First]{Abhishek Gupta}
\author[First]{Leena Vachhani}
\author[Fifth]{Omkar A. Ghag}

\address[First]{Indian Institute of Technology Bombay \\ E-mail: animeshsinghal.iitb@gmail.com, abhi.gupta@iitb.ac.in, leena.vachhani@iitb.ac.in}
\address[Second]{General Electric Research, Bangalore (E-mail:  sdsahil12@gmail.com)}
\address[Fifth]{Indian Institute of Technology, Bhubaneswar (E-mail: goa11@iitbbs.ac.in)}

\begin{abstract}                

A spherical robot has many practical advantages as the entire electronics are protected within a hull and can be carried easily by any Unmanned Aerial Vehicle (UAV). However, its use is limited due to finding mounts for sensors. Pendulum actuated spherical robot provides space for mounting sensors at the yoke. We study the non-linear dynamics of a pendulum-actuated spherical robot to analyze the dynamics of internal assembly (yoke) for mounting sensors. For such robots, we provide a coupled dynamic model that takes care of the relationship between forward and sideways motion. We further demonstrate the effects of wobbling and precession captured by our model when the bot is controlled to execute a turning maneuver while moving with a moderate forward velocity – a practical situation encountered by spherical robots moving in an indoor setting. A simulation setup based on the developed model provides visualization of the spherical robot motion.

\end{abstract}


\end{frontmatter}

\section{Introduction}
Spherical robots are mobile, ball-like robots. These bots consist of an outer shell, which rolls, with the actuation system and electrical circuitry enclosed inside this shell protected from extreme environments and external disturbances. When disturbed, these robots can reattain their equilibrium state due to their spherical shape. Due to these advantages, spherical robots are suitable for applications such as surveillance, reconnaissance, hazardous environment assessment, search, rescue, and planetary exploration as a rover. Many novel Unmanned Aerial Vehicle (UAV) operations can be facilitated by carrying a spherical robot and dropping it at a suitable location. \cite{antol2005new} discuss deployment concepts for such a surface exploration rover.



\begin{figure}[t!]
	\begin{subfigure}{0.2\textwidth}
		\centering
		\includegraphics[width=\linewidth]{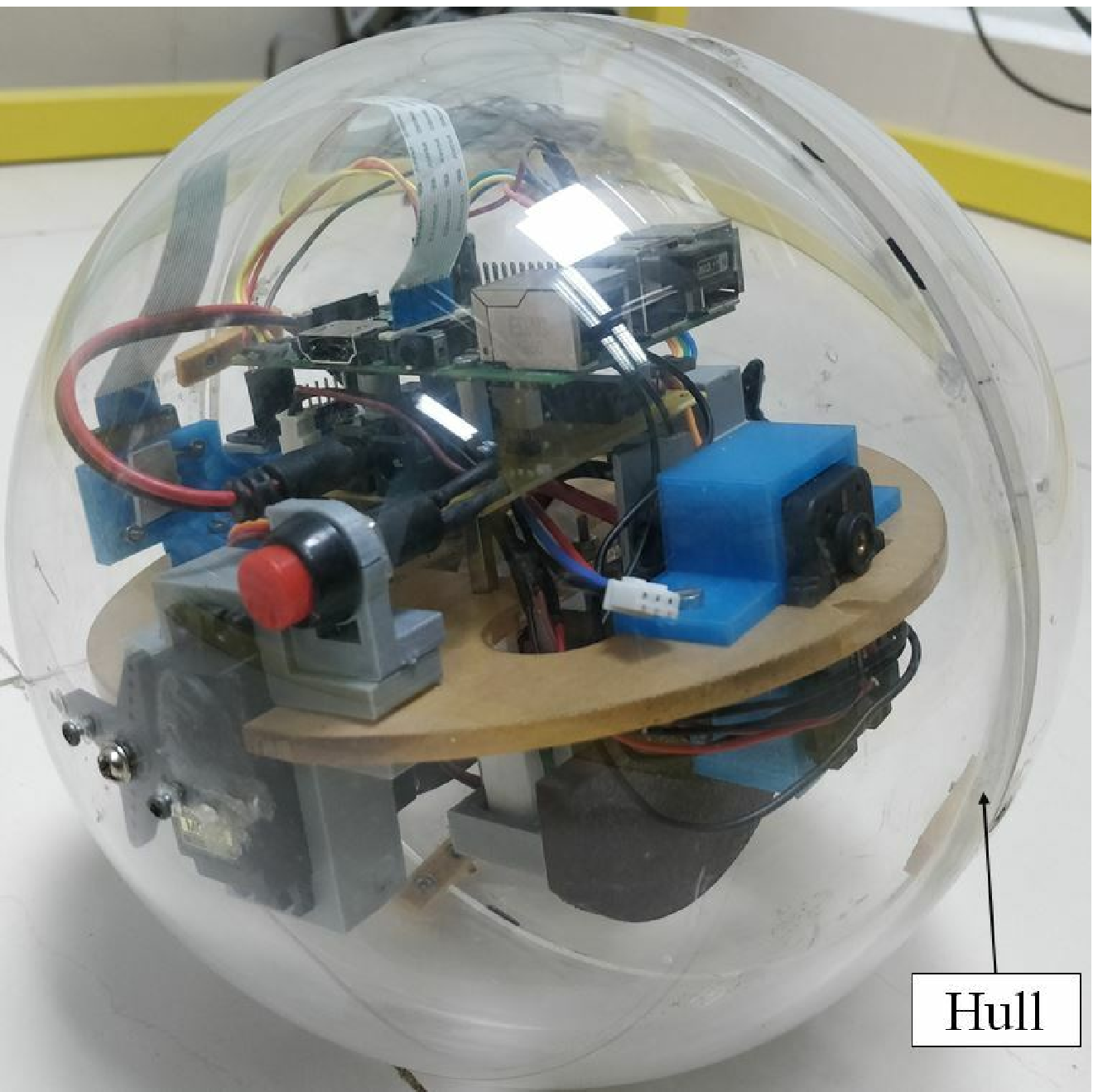}
		\label{fig:bot1}
		\caption{Assembled robot}
	\end{subfigure}
	\begin{subfigure}{0.25\textwidth}
        \centering
        \includegraphics[width=\linewidth]{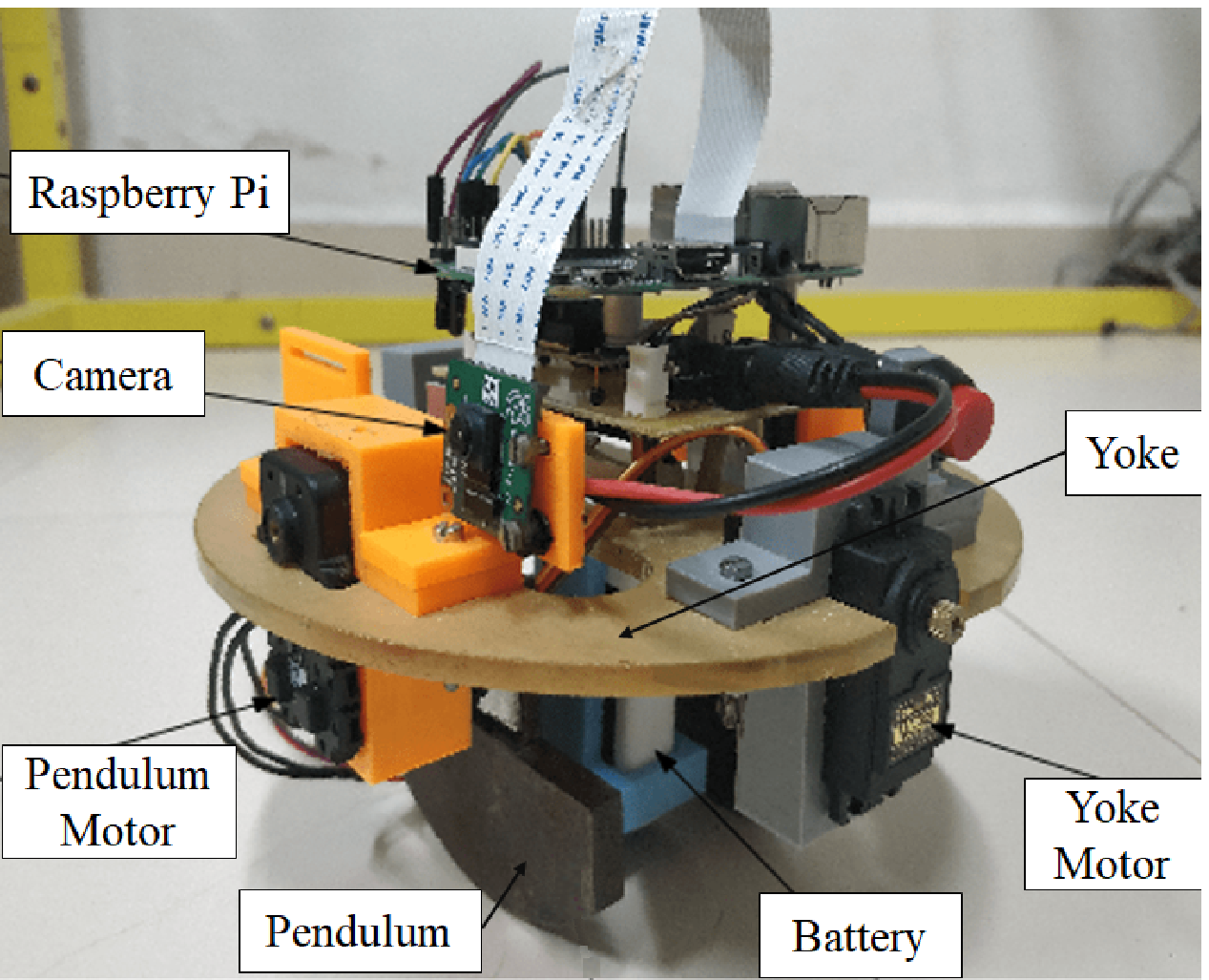} 
        \caption{Components of the spherical robot}
    \end{subfigure}
    \caption{The spherical robot \label{fig:bot}}
\end{figure}

In this paper, we consider a spherical bot with a yoke-pendulum arrangement for actuation, as shown in Fig. \ref{fig:bot}. The pendulum-actuated spherical robot has a major advantage of securing its equilibrium configuration while being dropped by a UAV. The driving mechanisms determine the robot's dexterity in terms of its ability to roll in multiple directions. A variety of such mechanisms have been developed, starting from the wheel drive by \cite{509415} to the pendulum actuated drives by  \cite{mahboubi2013design,Michaud,yoon2011spherical}. \citep{raura2017spherical, kalita2020dynamics} present a spherical planetary robot with a hopping mechanism to jump over rugged terrain. To address power supply limitations faced by internal-driven spherical robot to explore large region of planetary surfaces, \cite{li2011design} propose a wind-driven spherical robot with multiple shapes capable of rolling, bouncing, and flying by adjusting its external shape. Several other driving mechanisms can are discussed by \cite{crossley2006literature}, \cite{hajos2005overview} and \cite{tomik2012design}.

The forward and backward motion of the robot presented in this work is made possible by a motor attached to the yoke, which rotates the hull at the required speed. The pendulum is used to perturb the center of mass of the bot to provide steering motion. A camera, other sensors, and necessary electronics are mounted on the yoke.

Several modeling approaches have been proposed for representing the dynamics of the spherical robots in literature. First-order mathematical models of the spherical robots are based on the principle of the pure rolling constraint and conservation of angular momentum \citep{897794,844763}. The dynamics of a rolling sphere on a smooth surface have been modeled using the Lagrangian method and Euler angles \citep{rosen,ambloch}.

The dynamics of the sphere with a pendulum-based actuation have been studied in different works. \cite{liu2008family} derive a simplified dynamic model of driving ahead motion with respect to the drive motor torque as input. 
A unified model for all common decoupled dynamic models of pendulum-driven spherical robots have been developed \citep{ylikorpi2014unified}. \cite{li2012dynamic} present a model using Maggi’s equations without Lagrange’s multipliers from a view of the nonholonomic constraint. \cite{decoupled} model the pendulum and yoke-based spherical robot using a decoupling approach, in which the dynamics of the rolling and the steering motion are decoupled.  \cite{ylikorpi2014gyroscopic} develop the gyroscopic precision model with no-slip conditions and simulate circular motion trajectories for GimBall spherical robot. The models presented by \cite{897794}, \cite{decoupled} and \cite{cai2012path} demonstrate straight line and circular motion executed by spherical bots in simulation. \cite{lee2013design} present a probabilistic approach for generating a path between predefined start and destination locations considering the stochastic motions of spherical robots. 

Results presented in these works provide a bird-eye view of the motion. Still, they do not capture the small amplitude oscillations exhibited by the spherical robots while moving in simple paths like straight or a circle. \cite{8794742} demonstrate experimental results highlighting the shaky nature of pendulum-based spherical robots for various pendulum angles. Such oscillations become crucial when a robot houses a camera/sensors to capture its surroundings. Specifically, we are interested in the bot's wobbling or sideways fluctuations (perpendicular to the heading direction) as it leads to deviations in the bot's trajectory and unsteady video feedback from the mounted camera. Hence, the objective of this paper is to model the bot's dynamics and analyze the bot's behavior when it executes linear or circular motions. This analysis provided insights into the coupling within the robot's dynamics, which results in oscillatory behaviors. Such oscillations have been generally neglected in the literature due to their relatively smaller amplitude.

The paper is organized as follows. In the next section, we model the robot's dynamics using Euler angles and the Lagrange D'Alembert equation, considering the non-holonomic constraints. The following section shows the results of the simulation of the dynamics of the spherical bot using a numerical solver. These results are analyzed to highlight the problem of oscillations in motion. Conclusion and future work involved in this project are stated towards the end.


\section{Modeling and Dynamics}

 Fig. \ref{fig:bot} shows the pendulum actuated spherical bot. The bot comprises a hollow sphere called a hull, a platform housing all components called a yoke and a pendulum. The pendulum is mounted on the yoke at the geometric center of the hull through a motor. This motor rotates the pendulum with respect to the yoke resulting in sideways motion of the bot by providing a torque $T_p$. The yoke and the hull are connected through a motor mounted on the yoke. This motor rotates the hull with respect to the yoke resulting in a forward motion of the bot by providing a torque $T_s$. The center of mass of the hull and the yoke are at the geometric center of the hull, while the pendulum's center of mass is at a distance $R_p$ from the geometric center. We assume that the bot rolls without slipping and use Lagrange D'Alembert equations to find the equations of motion.


\subsection{Reference frames and Euler Angles}
Four reference frames are introduced as shown in Fig. \ref{fig:yxz}: a global inertial frame fixed to the ground ($\mathbf{G}$), and the three frames attached to the yoke ($\mathbf{Y}$), pendulum ($\mathbf{P}$) and the hull ($\mathbf{H}$) respectively with their origins at the geometric center of the hull. The yoke frame $\mathbf{Y}$ is defined to meet the following constraints: (1) The $z$-axis of the yoke is always aligned with the $z$-axis of the hull, and (2) The $x$-axis of the yoke frame lies in the global $XZ$-plane, i.e., it always remains parallel to the ground.

The orientation of the hull at any given instant is characterized by three $YXZ$ Euler angles $\phi$, $\theta$, and $\psi$. The orientation of the pendulum requires another angle $\beta$ due to an additional degree of freedom. The transformation between frames happens as follows :



\begin{figure}[!t]
\centering
\includegraphics[width=0.48\textwidth]{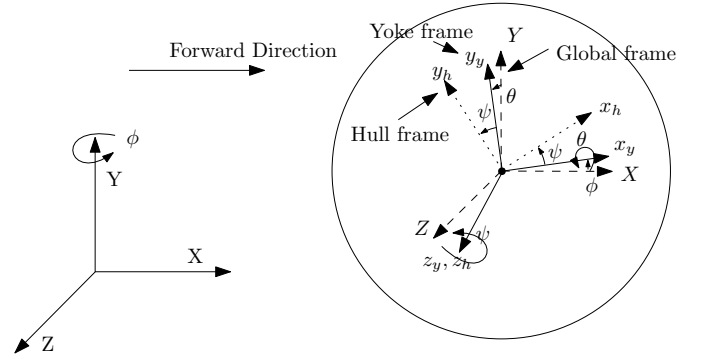}
\caption{Frames and Euler angles (YXZ)}
\label{fig:yxz}
\end{figure}


\begin{figure}[t!]
    \centering 
\begin{subfigure}{0.24\textwidth}
  \includegraphics[width=\linewidth]{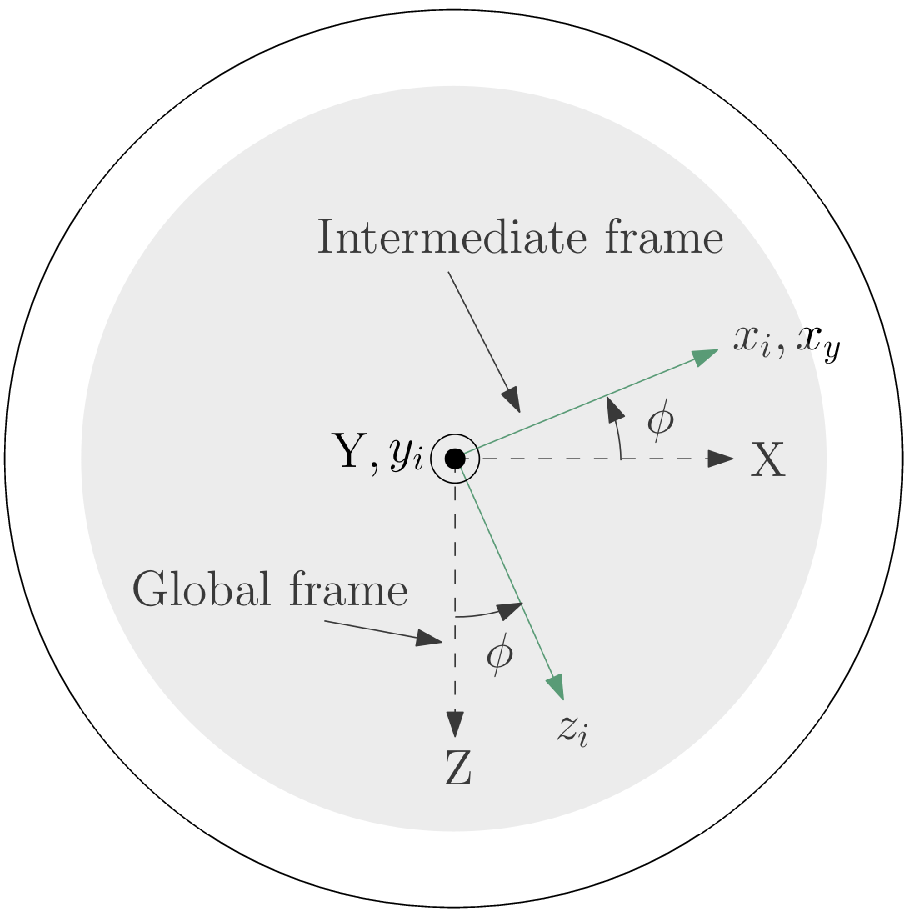}
  \caption{Rotation about Y axis (Top view)}
  \label{fig:fig3a_new}
\end{subfigure}\hfil 
\begin{subfigure}{0.24\textwidth}
  \includegraphics[width=\linewidth]{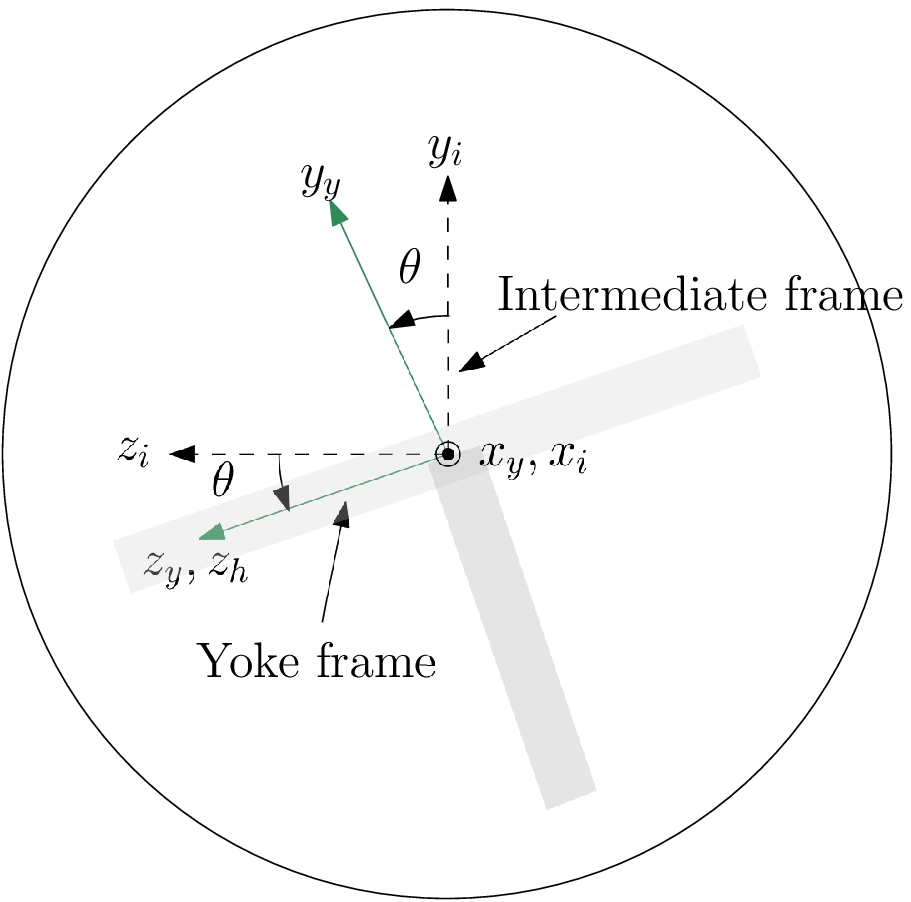}
  \caption{Rotation about $x_i$ axis (Front view)}
  \label{fig:fig3b_new}
\end{subfigure}
\begin{subfigure}{0.24\textwidth}
  \includegraphics[width=\linewidth]{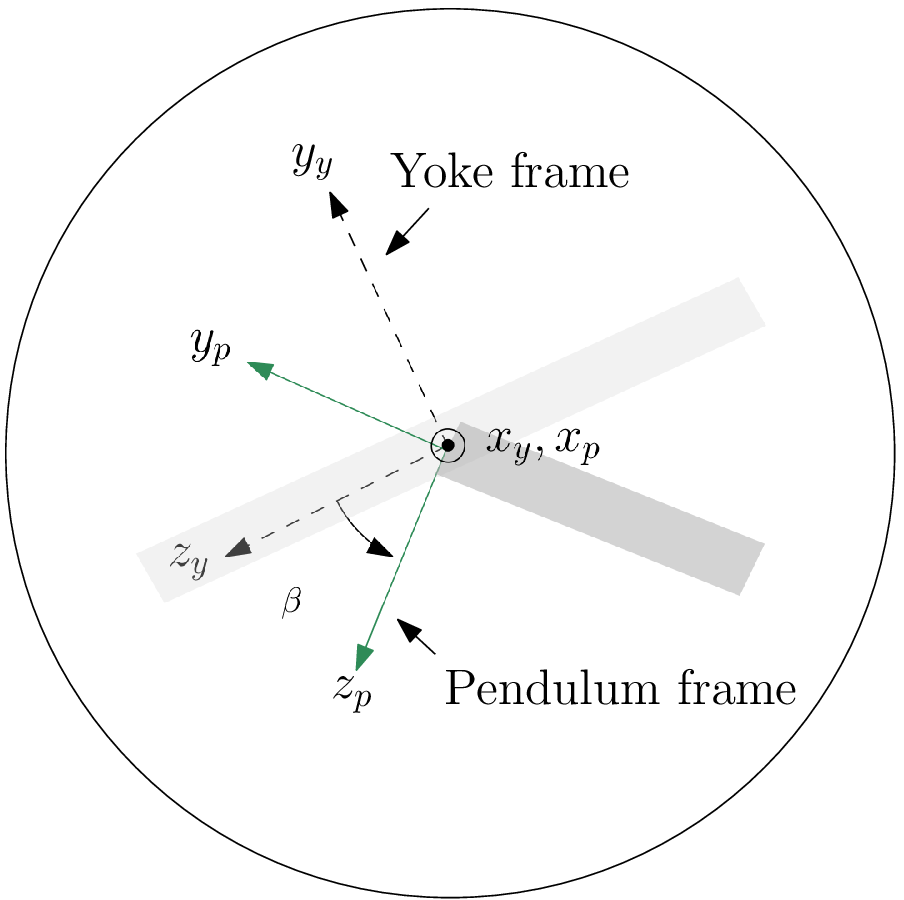}
  \caption{Rotation about $x_y$ axis (Front view)}
  \label{fig:fig3c_new}
\end{subfigure}
\begin{subfigure}{0.24\textwidth}
  \includegraphics[width=\linewidth]{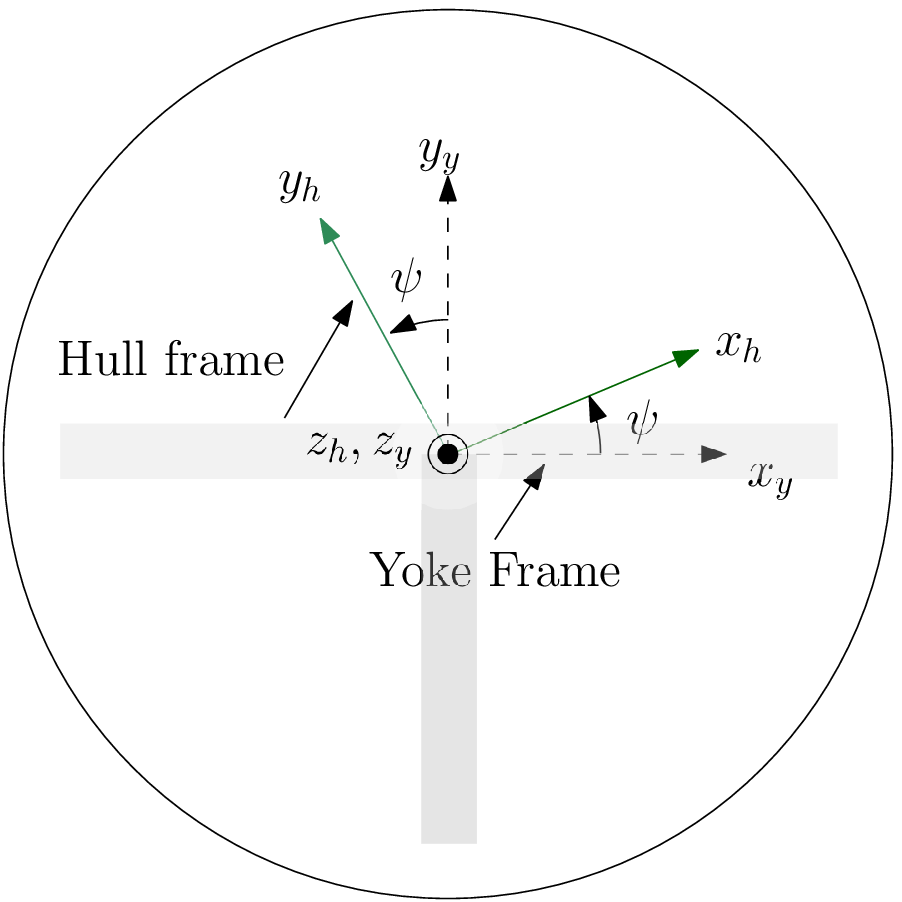}
  \caption{Rotation about $z_y$ axis (Side view)}
  \label{fig:fig3d_new}
\end{subfigure}
\caption{Steps for obtaining the frames}
\label{fig:fig3_new}
\end{figure}
\begin{itemize}
    \item Intermediate Frame ($\mathbf{I}$): $\mathbf{G}$ is rotated along its Y-axis by angle $\phi$ to obtain $\mathbf{I}$ as shown in Figure \ref{fig:fig3a_new}. $\phi$ represents the heading angle of the robot measured with respect to the global X-axis. This angle characterises the precession of the bot. 
    \item Yoke frame ($\mathbf{Y}$): $\mathbf{I}$ is rotated along its local $x$-axis $x_i$ by angle $\theta$ to obtain $\mathbf{Y}$ as shown in Figure \ref{fig:fig3b_new}. $\theta$ represents the lateral tilt of the robot perpendicular to the heading direction. This angle characterises the wobbling of the robot.
    \item Pendulum frame ($\mathbf{P}$): $\mathbf{Y}$ is rotated along its local $x$-axis $x_y$ by angle $\beta$ to obtain $\mathbf{P}$ as shown in Figure \ref{fig:fig3c_new}. This angle characterises the pendulum angle relative to the yoke. Note that the angle made by the pendulum with the vertical axis $Y$ is ($\beta+\theta$).    
    \item Hull frame ($\mathbf{H}$): $\mathbf{Y}$ is rotated along its $z$-axis $z_y$ by angle $\psi$ to obtain $\mathbf{H}$ as shown in Figure \ref{fig:fig3d_new}. $\psi$ represents the forward spin of the bot responsible for moving it forward or backwards. This angle characterises the forward rolling of the robot.
\end{itemize}

\subsection{Rotation Matrices}

The rotation matrix ${^A}R_B$ converts the velocities, angular velocities and other vectors from frame $B$ to frame $A$. Rotation matrices $R_X(\alpha)$, $R_Y(\alpha)$ and $R_Z(\alpha)$ are used to rotate vectors by an angle $\alpha$ about the x-, y-, or z-axes. The rotation matrices relating frames $\mathbf{Y}$, $\mathbf{P}$ and $\mathbf{H}$ to $\mathbf{G}$ can be written as, ${{^G}R_Y} = {R_Y(\phi)} {R_X(\theta)}$, ${{^G}R_P} = {R_Y(\phi)} {R_X(\theta)} {R_X(\beta)}$ and ${{^G}R_H} = {R_Y(\phi)} {R_X(\theta)} {R_Z(\psi)}$. 

\subsection{Angular Velocities}

Vector $\vec{{^Y}\omega_X}$ denotes the angular velocity of frame X represented in Y frame. The angular velocities  obtained by differentiating the rotation matrices relating that frame with the global frame are: 
\begin{equation}
\vec{{^Y}\omega_Y} = \begin{bmatrix}
\dot{\theta} && \dot{\phi}\,cos(\theta) && -\dot{\phi}\,sin(\theta)
\end{bmatrix}^\intercal{}
\end{equation}
\begin{equation}
\vec{{^P}\omega_P} = \begin{bmatrix}
\dot{\beta}+\dot{\theta} &&
\dot{\phi}\,\cos(\beta+\theta) &&
-\dot{\phi}\,\sin(\beta+\theta)
\end{bmatrix}^\intercal{} 
\end{equation}
\begin{equation}
\vec{{^H}\omega_H} = \begin{bmatrix}
\dot{\theta}\,cos(\psi)\,+\,\dot{\phi}\,cos(\theta)\,sin(\psi) \\
\dot{\phi}\,cos(\psi)\,cos(\theta)\,-\,\dot{\theta}\,sin(\psi) \\
\dot{\psi}\,-\,\dot{\phi}\,sin(\theta)
\end{bmatrix}
\end{equation}

\subsection{Linear Velocities}

Position vector of a point P in coordinate frame $B(oxyz)$ expressed in the global frame $G(OXYZ)$ is denoted by $\vec{{}^{G}_{B}r_{P}}$. The position vector corresponding to the center of mass of the yoke and the hull are
\begin{equation}
\vec{{}^{G}_{G}r_{Y_c}} = \vec{{}^{G}_{G}r_{H_c}} = 
\begin{bmatrix}
X & R_s & Z
\end{bmatrix}^\intercal{}
\label{rHull}
\end{equation}
where $X$ and $Z$ are the co-ordinates of the hull centre along the $x$ axis and $z$ axis of the global frame respectively and $R_s$ is the radius of the sphere. Note that the origins of frames $\mathbf{Y, P, H}$ coincide at the center of the bot. The position vector of Pendulum's centre of mass in global frame expressed in global frame is given by
\begin{equation}
\vec{{}^{G}_{G}r_{P_c}} = \vec{{}^{G}_{P}r_{P_c}} + \vec{{}^{G}_{G}r_{P_o}} 
\label{rPendulum}
\end{equation}
\begin{equation}
\vec{{}^{G}_{P}r_{P_c}} = {^G}R_P \vec{{}^{P}_{P}r_{P_c}}   
\text{ and }\vec{{}^{P}_{P}r_{P_c}} = 
\begin{bmatrix}
0 & -R_p & 0
\end{bmatrix}^\intercal{}
\end{equation}
\begin{equation}
\vec{{}^{G}_{G}r_{P_o}} = \begin{bmatrix}
X & R_s & Z
\end{bmatrix}^\intercal{}
\end{equation}

where $P_c$ stands for pendulum's centre of mass, $P_o$ stands for origin of pendulum frame and $R_p$ is the distance between $P_c$ and $P_o$. 

Differentiating the position vectors, we obtain the velocities of the center of mass of the hull, yoke and the pendulum as
\begin{equation}
\vec{{^G}v_{H_c}} = \vec{{^G}v_{Y_c}} =
\begin{bmatrix}
\dot{X} & 0 & \dot{Z}
\end{bmatrix}^\intercal{}
\label{linVelHull}
\end{equation}
\begin{equation}
\vec{{^G}v_{P_c}} = 
\begin{bmatrix}
\dot{X} \\ 0 \\ \dot{Z}
\end{bmatrix} + \vec{{^G}\omega_P} \times {^G}R_P \begin{bmatrix}
0 \\ -R_p \\ 0
\end{bmatrix} 
\end{equation}

\subsection{Energies}
The Kinetic Energy of the system can now be written as
\begin{multline}
K = {\frac{1}{2}}(m_H \|\vec{{^G}v_{H_c}}\|^2 +
m_Y \|\vec{{^G}v_{Y_c}}\|^2 + 
m_P \|\vec{{^G}v_{P_c}}\|^2) + \\
{\frac{1}{2}}(\vec{{^H}\omega_H}^\intercal {^H}I_H \vec{{^H}\omega_H} + 
\vec{{^Y}\omega_Y}^\intercal {^Y}I_Y \vec{{^Y}\omega_Y} + 
\vec{{^P}\omega_P}^\intercal {^P}I_P \vec{{^P}\omega_P})
\end{multline}
where $m_X$ denoted the mass of body X and ${^X}I_X$ denotes the mass moment of inertia matrix of body X calculated in its own frame of reference and can be approximated as  ${^H}I_H = {\frac{2}{3}}*diag(m_H R_s^2, m_H R_s^2, m_H R_s^2)$, ${^Y}I_Y = {\frac{1}{4}} * diag(m_Y R_s^2, 2 m_Y R_s^2, m_Y R_s^2)$ and ${^P}I_P = {\frac{1}{3}} * diag(m_P R_p^2, 0,m_P R_p^2)$.

The Potential Energy of the system is given by
\begin{equation}
V = m_P g (\vec{{}^{G}_{P}r_{P_c}}\cdot{[0\text{ }1\text{ }0]}^\intercal) + 
m_Y g (\vec{{}^{G}_{Y}r_{Y_c}}\cdot{[0\text{ }1\text{ }0]}^\intercal)
\end{equation}
where the datum point for potential energy is chosen as the geometric centre of the sphere. 

\subsection{Non-holonomic Constraints}

The constraint of rolling without slipping is non-holonomic and is given by
\begin{equation}
\vec{{^G}v_H} = \vec{{^G}\omega_H} \times \vec{{^G}r_H}
\end{equation}
where $\vec{{^G}r_H} = \begin{bmatrix} 0 & R_s & 0
\end{bmatrix}^\intercal$ denotes the vector (written in global frame) between the stationary point on the hull which is in contact with the ground and the sphere centre.  The above equation can be simplified further as
\begin{equation}
\dot{X} = R_s (\dot{\theta} sin(\phi) - \dot{\psi}cos(\phi)cos(\theta))
\label{xdot_eq}
\end{equation}
\begin{equation}
\dot{Z} = R_s (\dot{\theta} cos(\phi) + \dot{\psi}sin(\phi)cos(\theta))
\label{zdot_eq}
\end{equation}

Note that equations \ref{xdot_eq} and \ref{zdot_eq} cannot be substituted directly in equation \ref{linVelHull} and solved in Lagrangian because an important assumption while deriving the Lagragian equation is that the generalized co-ordinates are independent of each other \citep{ambloch}. Therefore the constraint equations are written in the form
\begin{equation}
\vec{a_1}^\intercal\cdot\dot{\vec{q}} = 0 \text{ , } \vec{a_2}^\intercal\cdot\dot{\vec{q}} = 0
\end{equation}
where  
\begin{equation}
\dot{\vec{q}} = \begin{bmatrix}
\dot{X} & \dot{Z} & \dot{\phi} & \dot{\theta} & \dot{\psi} & \dot{\beta}
\end{bmatrix}^\intercal \\
\end{equation}
is the vector of time derivative of generalized co-ordinates and $\vec{a_1}$ and $\vec{a_2}$ are
\begin{equation}
\vec{a_1} = \begin{bmatrix}
1 & 0 & 0 & -R_s sin(\phi) & R_s cos(\phi)cos(\theta) & 0
\end{bmatrix}^\intercal \\
\label{q_a1_vec}
\end{equation}
\begin{equation}
\vec{a_2} = \begin{bmatrix}
0 & 1 & 0 & -R_s cos(\phi) & -R_s sin(\phi)cos(\theta) & 0
\end{bmatrix}^\intercal \\
\label{q_a2_vec}
\end{equation}

\subsection{Lagrange D'Alembert equations}

The Lagrangian of the system is given by $L = K - V$. 
The Lagrange D'Alembert equations for non-holonomic systems are given by 
\begin{equation}
\frac{d}{dt}(\frac{\partial L}{\partial \dot{q}}) - \frac{\partial L}{\partial q} = Q + \lambda_1 a_1 + \lambda_2 a_2
\end{equation}
where Q is the generalized forces, $\lambda_1$ and $\lambda_2$ are Lagrangian parameters and $a_1$ and $a_2$ are vectors are given by equations \ref{q_a1_vec}, \ref{q_a2_vec}. As $T_s$ is the torque applied for forward motion and $T_p$ is the torque applied on the pendulum, Q is given by $[0\text{ }0\text{ }0\text{ }0\text{ }T_s\text{ }T_p]^\intercal$. Rewriting the equation for each generalized co-ordinate,
\begin{equation}
\frac{d}{dt}(\frac{\partial L}{\partial \dot{X}}) = \lambda_1
\label{eqX}
\end{equation}
\begin{equation}
\frac{d}{dt}(\frac{\partial L}{\partial \dot{Z}}) = \lambda_2
\label{eqZ}
\end{equation}
\begin{equation}
\frac{d}{dt}(\frac{\partial L}{\partial \dot{\phi}}) - \frac{\partial L}{\partial \phi} = 0
\label{eqPhi}
\end{equation}
\begin{equation}
\frac{d}{dt}(\frac{\partial L}{\partial \dot{\theta}}) - \frac{\partial L}{\partial \theta} = -R_s sin(\phi)\lambda_1 -R_s cos(\phi)\lambda_2
\label{eqTheta}
\end{equation}
\begin{multline}
\frac{d}{dt}(\frac{\partial L}{\partial \dot{\psi}}) - \frac{\partial L}{\partial \psi} = R_s cos(\phi)cos(\theta)\lambda_1 - \\ R_s sin(\phi)cos(\theta) \lambda_2 + T_s
\label{eqPsi}
\end{multline}
\begin{equation}
\frac{d}{dt}(\frac{\partial L}{\partial \dot{\beta}}) - \frac{\partial L}{\partial \beta} = T_p
\label{eqBeta}
\end{equation}

Substituting equations \ref{eqX}-\ref{eqZ} into \ref{eqTheta}-\ref{eqPsi} along with the equations \ref{xdot_eq} and \ref{zdot_eq} gives the dynamic model.  
Using chain rule of differentiation, the equations can be written in a consolidated form as
\begin{equation}
M(\mathbf{x},t)\dot{\mathbf{x}} = f(\mathbf{x},t)
\label{dynamics_final}
\end{equation}
where $\mathbf{x}$ is the state vector given by
\begin{equation}
{\mathbf{x}} = 
\begin{bmatrix}
\phi & \theta & \psi & X & Z & \dot{\phi} & \dot{\theta} & \dot{\psi} & \dot{X} & \dot{Z} \textbf{ }\textbf{ } \beta \textbf{ }\textbf{ } \dot{\beta}
\end{bmatrix}^\intercal
\label{state_vector}
\end{equation}


The dynamics of spherical robot given by equation \ref{dynamics_final} are simulated using the in-built MATLAB solver ODE15s. 


\section{Analysis of Wobbling and Precession}

In this section, we analyze the system response exhibited by the bot for a turning maneuver using the dynamic model given by equation \ref{dynamics_final}. The bot is controlled via forward rolling torque $T_s$ and pendulum torque $T_p$ to command a specific forward velocity $\dot{\psi}$ and pendulum angle $\beta$ with respect to yoke respectively. 
We design a pendulum controller using $T_p$ consisting of a feed-forward term to counter-act gravity torque and a feed-back term comprising a PD controller with high gains to quickly settle $\beta$ to a desired value. We also design a speed controller using $T_s$ having a feedback term with a proportional controller with high gain so that the bot starts moving with a desired forward velocity $\dot{\psi}$. High gains in both controllers help to keep the focus on the steady-state response instead of the transients leading to it.

\begin{figure}
    \centering 
\begin{subfigure}{0.22\textwidth}
  \includegraphics[width=0.85\linewidth, trim = 1cm 0 1cm 0]{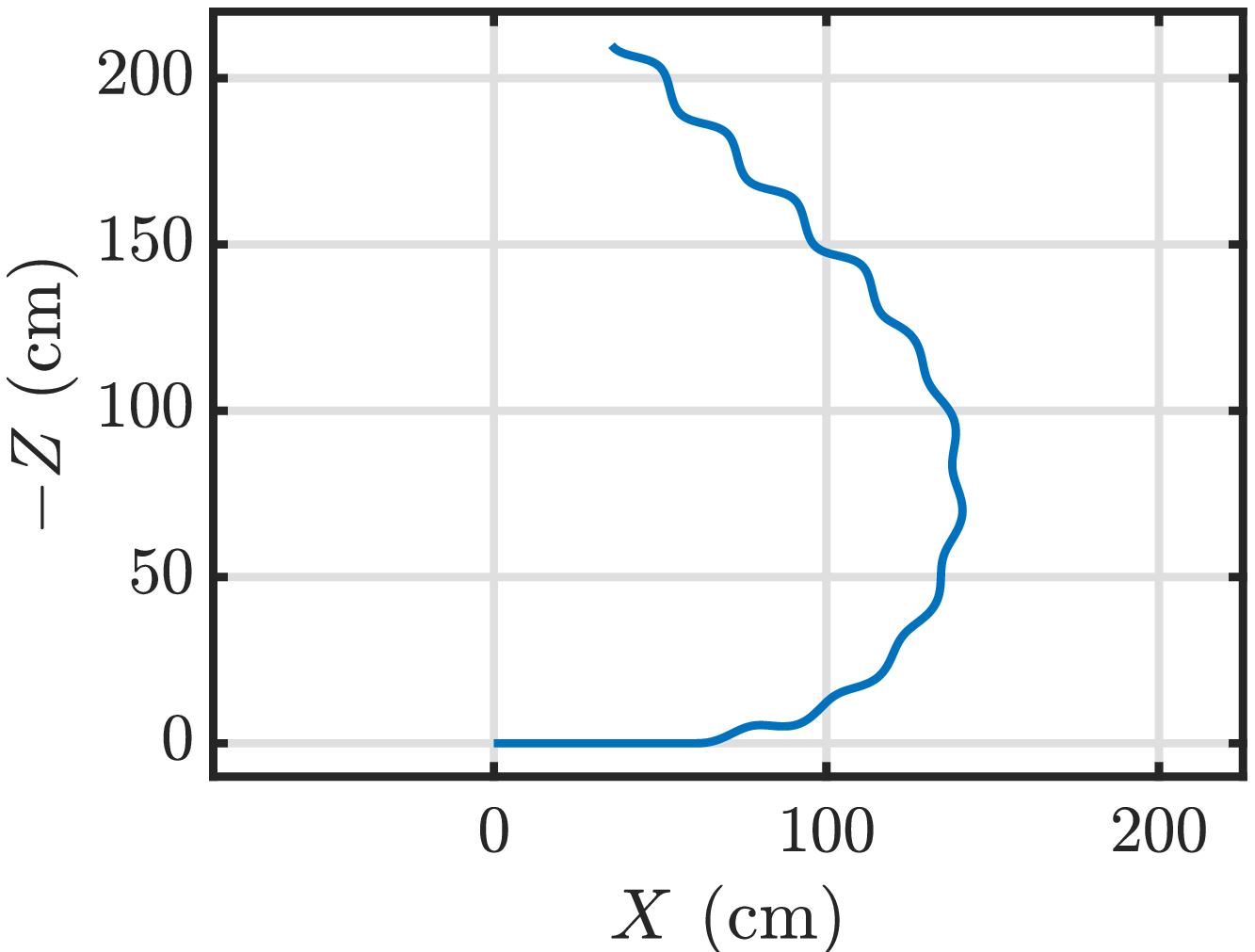}
  \caption{\centering{Bot's trajectory}}
  \label{fig:TurningTrajectory}
\end{subfigure}\hfil 
\begin{subfigure}{0.22\textwidth}
  \includegraphics[width=0.85\linewidth, trim = 1cm 0 1cm 0]{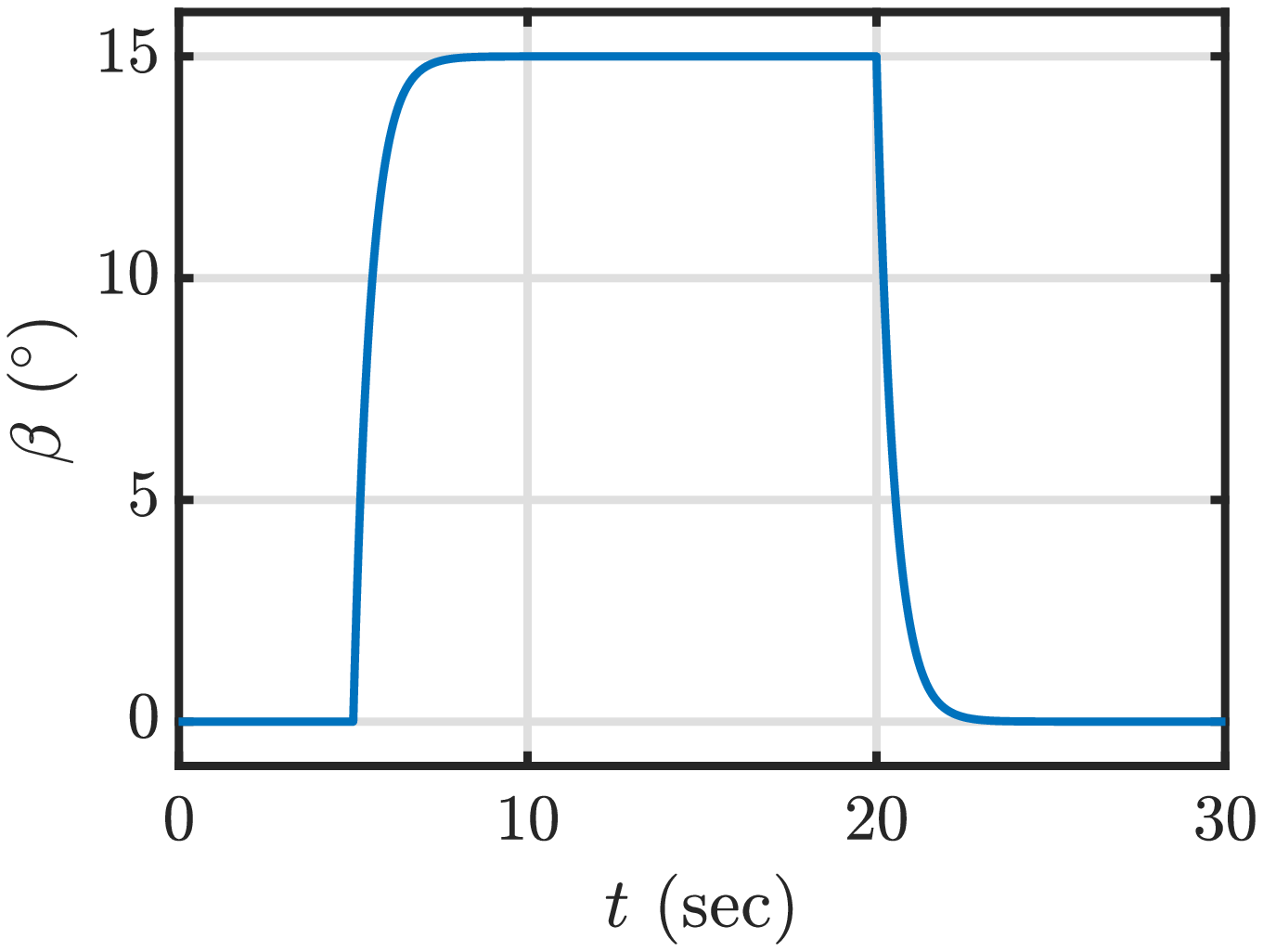}
  \caption{\centering{Pendulum angle vs. time}}
  \label{fig:TurningBeta}
\end{subfigure}

\medskip
\begin{subfigure}{0.22\textwidth}
  \includegraphics[width=0.85\linewidth, trim = 1cm 0 1cm 0]{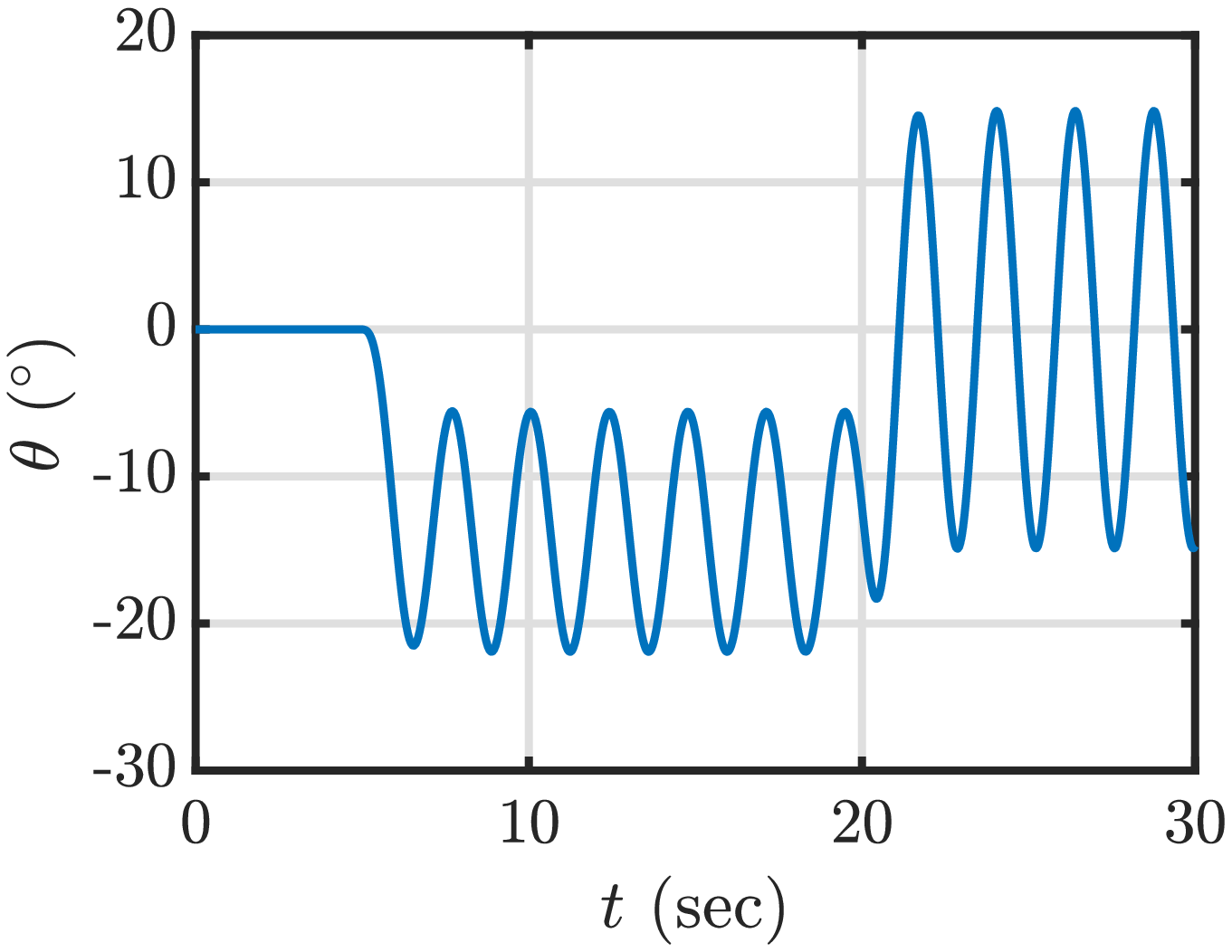}
  \caption{\centering{Wobbling vs. time }}
  \label{fig:TurningWobble}
\end{subfigure}\hfil 
\begin{subfigure}{0.22\textwidth}
  \includegraphics[width=0.85\linewidth, trim = 1cm 0 1cm 0]{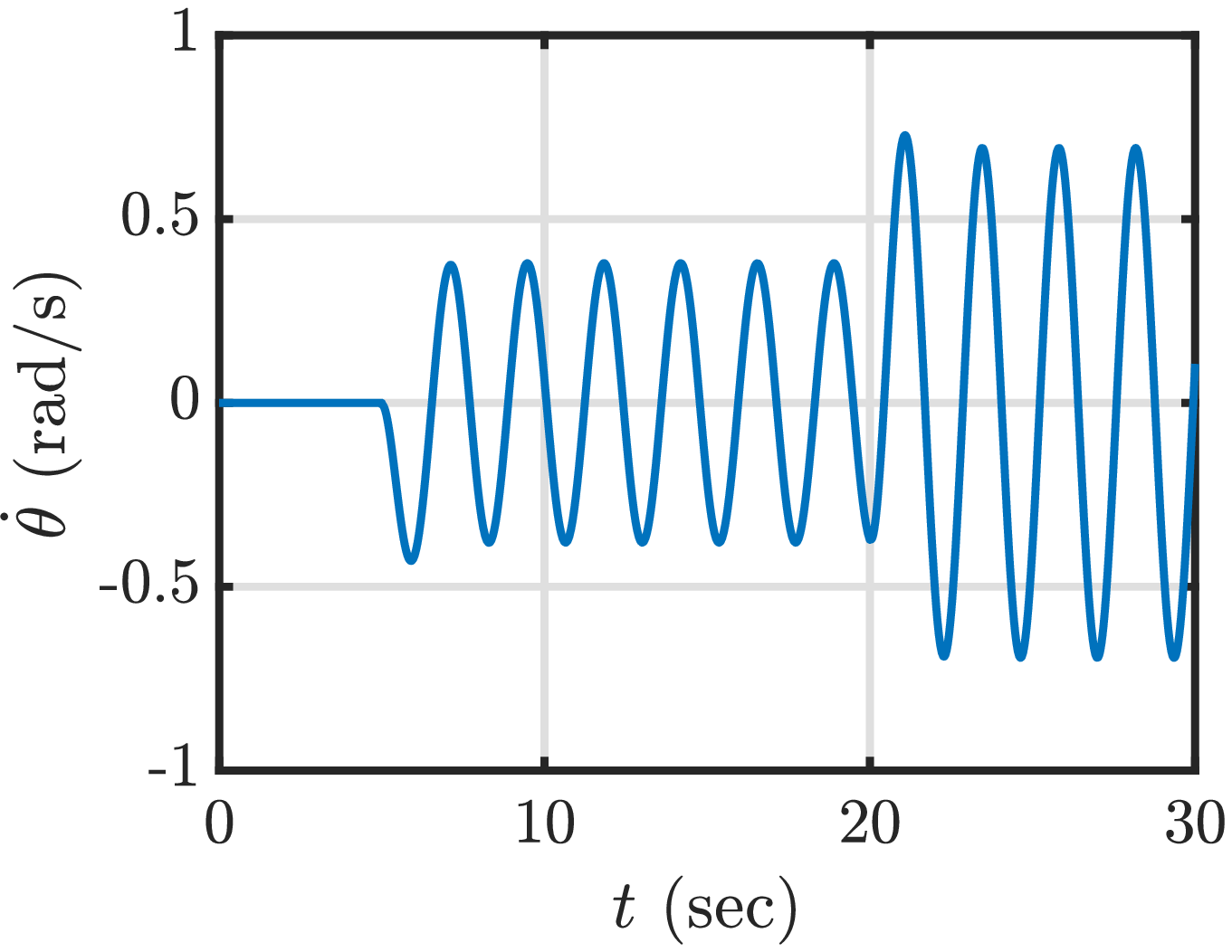}
  \caption{\centering{Wobbling rate vs. time}}
  \label{fig:TurningWobbleRate}
\end{subfigure}

\medskip
\begin{subfigure}{0.22\textwidth}
  \includegraphics[width=0.85\linewidth, trim = 1cm 0 1cm 0]{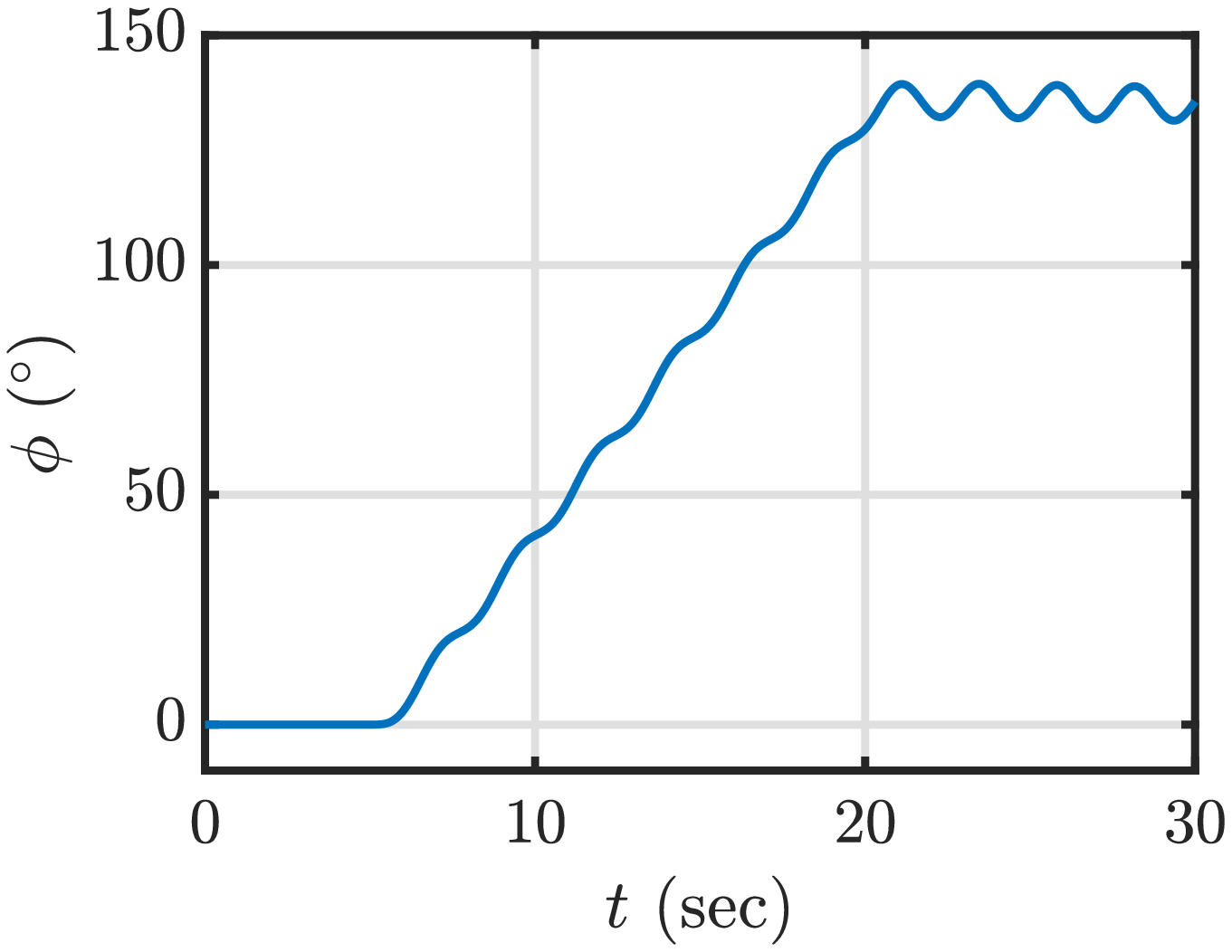}
  \caption{\centering{Precession vs. time}}
  \label{fig:TurningHeading}
\end{subfigure}\hfil 
\begin{subfigure}{0.22\textwidth}
  \includegraphics[width=0.85\linewidth, trim = 1cm 0 1cm 0]{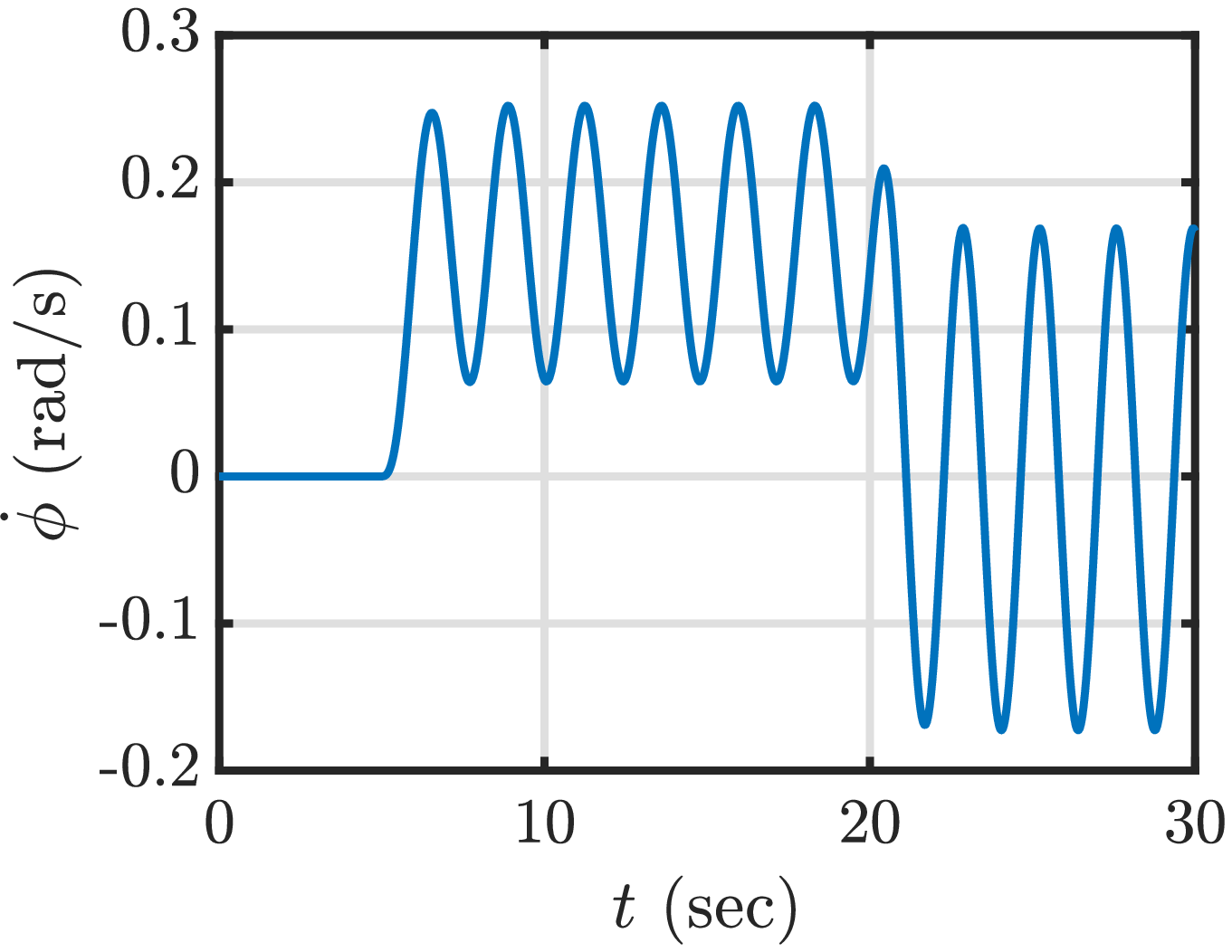}
  \caption{\centering{Precession rate vs. time}}
  \label{fig:TurningHeadingRate}
\end{subfigure}

\medskip
\begin{subfigure}{0.22\textwidth}
  \includegraphics[width=0.85\linewidth, trim = 1cm 0 1cm 0]{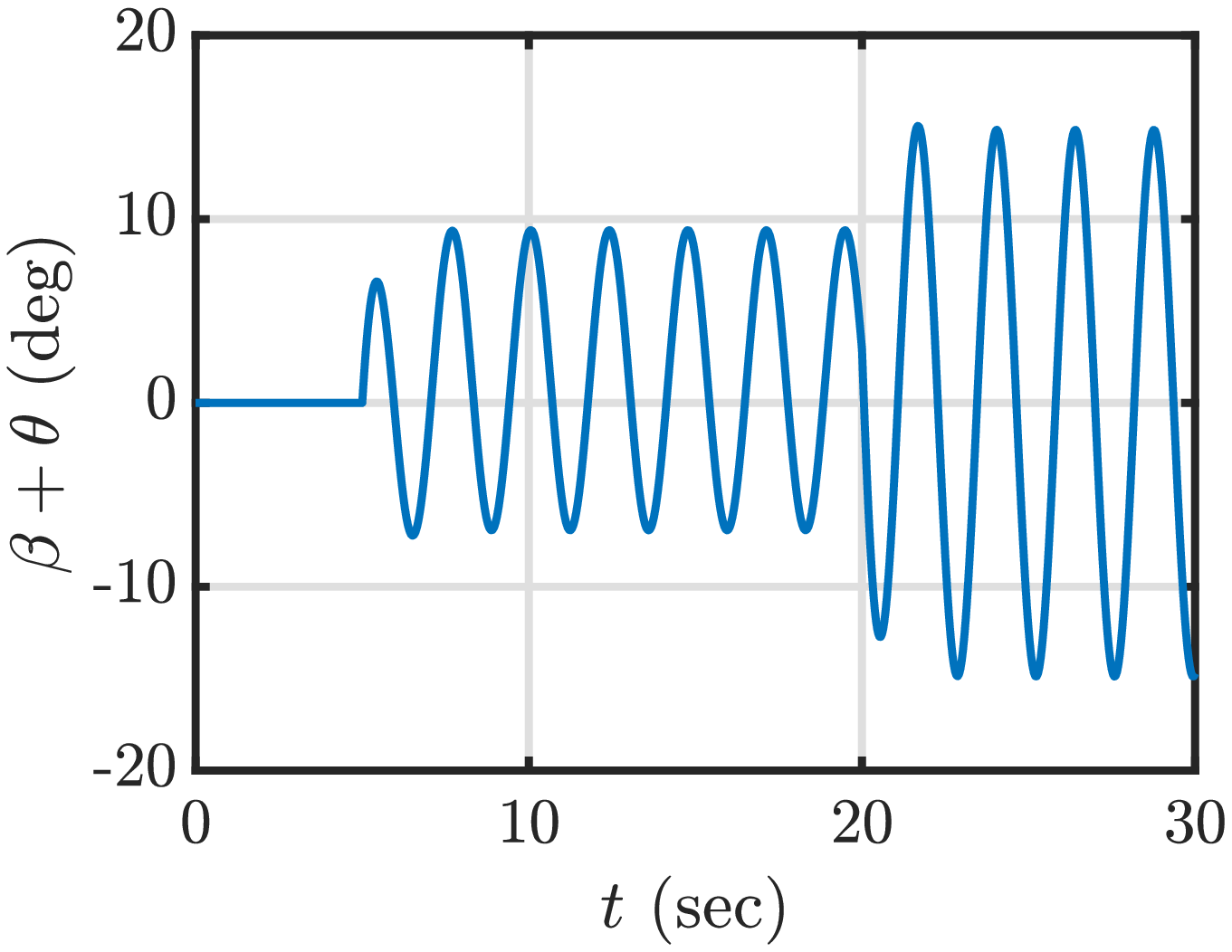}
  \caption{\centering{Pendulum angle wrt vertical vs. time}}
  \label{fig:TurningPendulum}
\end{subfigure}\hfil 
\begin{subfigure}{0.22\textwidth}
  \includegraphics[width=0.85\linewidth, trim = 1cm 0 1cm 0]{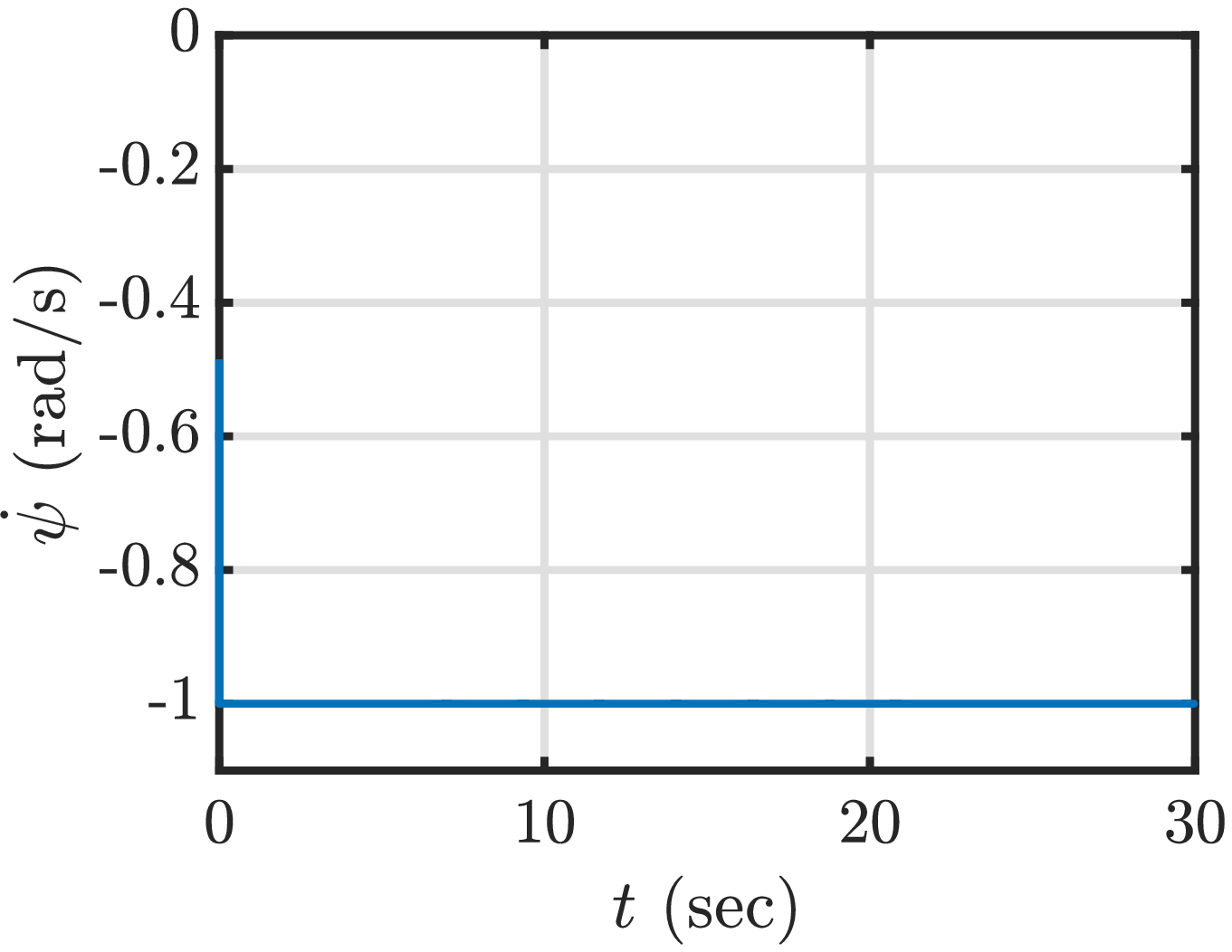}
  \caption{\centering{Forward rolling angular speed vs. time}}
  \label{fig:TurningPsid}
\end{subfigure}

\caption{Analysis for Turning motion} 
\label{fig:TurningStateResponse}
\end{figure}

We present a case where the bot starts moving in a straight line. Then the pendulum is displaced to an angle that introduces wobbling, as shown in figure \ref{fig:TurningWobble} in the system while moving the bot in an arc. We finally bring back the pendulum to its initial configuration. This case represents a turning motion of the bot as shown in figure \ref{fig:TurningTrajectory}.

Based on the steady-state analysis discussed in the literature, we can expect the bot to go straight, turn in a circle and finally resume moving in a straight line when it executes a turning motion introduced above. In reality, the spherical bot wobbles along its way as observed in the experiments\footnote{The videos at \url{https://tiny.cc/SphericalBotMotion} show the turning maneuver of a spherical bot in reality vs. 3D simulations generated using the dynamic model presented in this paper}.

The system response presented in figure \ref{fig:TurningStateResponse} demonstrates that our model 
qualitatively captures 
wobbling and precession exhibited by spherical robots. Our mathematical model characterizes wobbling (fig \ref{fig:TurningWobble}) by the Euler angle $\theta$ about the yoke's x-axis and precession (fig \ref{fig:TurningHeading}) by the Euler angle $\phi$ about the global y-axis. Wobbling rate (fig \ref{fig:TurningWobbleRate}) and precession rates (fig \ref{fig:TurningWobbleRate}) are thus given by $\dot\theta$ and $\dot\phi$ respectively. 

In the following subsections, we separately analyze the initial linear, circular, and final linear motions that constitute the bot's turning maneuver. In this setup, a fixed value of $\beta$ implies that the pendulum doesn't oscillate relative to the yoke. Note that the angle made by the pendulum with the vertical ($\theta+\beta$) determines the gravitational torque $\vec{\tau}$ experienced by the bot that acts along the yoke's x-axis. This torque is one of the main drivers behind the precession of the bot. A 3D visualization\footnote{See footnote 1} of the simulation demonstrate the pose of the spherical robot using generalised coordinates $\phi$, $\theta$, $\psi$, $\beta$, $X$ and $Z$.

\subsection{Initial straight line motion: t < 5 sec}
The bot is disturbed from its initial equilibrium configuration by imparting a forward velocity $\dot\psi$ of 1 rad/s using $T_s$. The bot quickly attains that value, as shown in figure \ref{fig:TurningPsid}. The bot continues to move at that speed in a straight line for 5 seconds, as shown in figure \ref{fig:TurningTrajectory}. $\dot{X}_0$ also stays non-zero due to pure rolling constraints. The bot does not wobble or precesses in this period, as shown in figures \ref{fig:TurningWobble} and \ref{fig:TurningHeading}. The pendulum also stays aligned with the vertical (ref. fig \ref{fig:TurningPendulum}.), and the pendulum stays perpendicular to the yoke as $\beta$ remains zero initially (ref. fig \ref{fig:TurningBeta}).

\subsection{Circular motion: 5 sec < t < 20 sec}
The bot's pendulum angle relative to the yoke $\beta$ is changed to 15$^{\circ}$ after moving in a straight line for 5 seconds by varying $T_p$. $\beta$ is further kept fixed for some duration as shown in (ref. fig \ref{fig:TurningBeta}). Bot's forward velocity $\dot\psi$ is left unchanged (ref. fig \ref{fig:TurningPsid}). As a result, the robot starts tracing a circular trajectory as shown in fig \ref{fig:TurningTrajectory}.

\subsubsection{Wobbling}
Figure \ref{fig:TurningWobble} shows the evolution of $\theta$ with time. The bot experiences wobbling because the pendulum motor disturbs the bot's stable configuration by changing the pendulum angle. To satisfy the pure rolling constraints, the oscillations in $\theta$ give rise to lateral displacements perpendicular to the heading direction of the bot when it is moving in a circle.

\subsubsection{Precession}
Figure \ref{fig:TurningHeading} shows the evolution of $\phi$ with time. It keeps increasing with time and also has an oscillatory component. The torque $\vec{\tau}$ generated due to the pendulum's weight contributes to both wobbling and precession. At high forward speed $\dot\psi$, precession dominates over wobbling as $\vec{\tau}$ is approximately perpendicular to the angular momentum $\vec{\mathbf{L}}$ that acts along the z-axis of the yoke and exists due to non-zero $\dot\psi$. At low forward speed $\dot\psi$, wobbling dominates over precession as $\vec{\tau}$ contributes to the change in $\theta$. In this case, bot's $\vec{\mathbf{L}}$ turns out to be either approximately zero or is parallel to yoke's x-axis if $\dot\theta$ is non-zero. In the simulation performed here, a moderate forward speed $\dot\psi$ (ref. fig \ref{fig:TurningPsid}) was simulated to observe a combination of both behaviors. 

The linearly increasing behavior of $\phi$ can be attributed to the non-zero average value of ($\beta+\theta$) as seen in figure \ref{fig:TurningPendulum}. This ensures that the mean position of the angle made by the pendulum with the vertical is non-zero. Due to this, $\vec{\tau}$ stays approximately unchanged. As a result, the bot precesses in an arc of approximately constant radius. 
The oscillations in $\phi$ occur due to the sinusoidal nature of ($\beta+\theta$) about a non-zero mean position, as seen in figure \ref{fig:TurningPendulum}. This implies that the pendulum-yoke-hull assembly oscillates about its mean position giving rise to a minor fluctuation in $\vec{\tau}$. This leads to variation in the bot's precession.

\subsection{Final straight line motion: t > 20 sec}
The bot's pendulum is finally made perpendicular to the yoke by changing $\beta$ back to 0$^{\circ}$ after by varying $T_p$. $\beta$'s position is maintained for the rest of the simulation time (ref. fig \ref{fig:TurningBeta}). Bot's forward velocity $\dot\psi$ is left unchanged (ref. fig \ref{fig:TurningPsid}). As a result, the robot moves in a straight line tangentially to the circular arc it executed, as shown in fig \ref{fig:TurningTrajectory}.

\subsubsection{Wobbling}
Figure \ref{fig:TurningWobble} shows the evolution of $\theta$ with time. Even though the pendulum becomes perpendicular to the yoke, the bot oscillates as it has already been displaced from its equilibrium configuration. Note that the wobbling amplitude is greater than when it was moving in a circle. To satisfy the pure rolling constraints, the wobbling nature of the bot also induced a lateral motion perpendicular to its heading direction, as observed in Figure \ref{fig:TurningTrajectory}.      

\subsubsection{Precession}
Figure \ref{fig:TurningHeading} shows the evolution of $\phi$ with time. The net change in $\phi$ over time averages out to be zero because of its oscillatory behavior. In this simulation, the angular momentum $\vec{\mathbf{L}}$ is dominated by $\dot\psi_0$ in comparison to $\dot\theta_0$ by an order of 10 (refer figure \ref{fig:TurningPsid} and \ref{fig:TurningWobbleRate}). Hence, the gravitational torque $\vec{\tau}$ due to the pendulum is approximately perpendicular to $\vec{\mathbf{L}}$. This results in the precession of the bot. But, as the value of ($\beta+\theta$) averages out to 0 over time, as seen in Figure \ref{fig:TurningPendulum}, the torque vector keeps flipping its direction resulting in a net zero change in precession over time.

\section{Conclusion and future work}

This paper investigates the option of building novel applications combining UAVs and spherical robots. In this regard, we present a dynamic model of pendulum actuated spherical bot that accounts for the coupling between forward and sideways motion. 
Our model successfully captures the oscillations of internal assembly. 
Two important classes of oscillations, i.e., wobbling and precession, were defined, analyzed, and explained in this context of spherical bots. The visualization tool (simulation set-up) for spherical robot motion further aids in qualitatively comparing the system response with actual robot behavior. Modeling these oscillations is essential to open up the scope to design a controller in the future for a stabilized yoke to mount sensors on the spherical robot. 
Controller for heading correction, trajectory correction, wobbling control, etc., can facilitate the uses of the spherical robot. E.g., $\phi$ and $\theta$ as output would give a heading and a wobbling controller, respectively. 

\bibliography{ifacconf}

\begin{thebibliography}{23}
\providecommand{\natexlab}[1]{#1}
\providecommand{\url}[1]{\texttt{#1}}
\providecommand{\urlprefix}{URL }
\expandafter\ifx\csname urlstyle\endcsname\relax
  \providecommand{\doi}[1]{doi:\discretionary{}{}{}#1}\else
  \providecommand{\doi}{doi:\discretionary{}{}{}\begingroup
  \urlstyle{rm}\Url}\fi

\bibitem[{Antol(2005)}]{antol2005new}
Antol, J. (2005).
\newblock A new vehicle for planetary surface exploration-the mars tumbleweed
  rover.
\newblock In \emph{1st Space Exploration Conference: Continuing the Voyage of
  Discovery}, 2520.

\bibitem[{Bhattacharya and Agrawal(2000{\natexlab{a}})}]{844763}
Bhattacharya, S. and Agrawal, S.K. (2000{\natexlab{a}}).
\newblock Design, experiments and motion planning of a spherical rolling robot.
\newblock In \emph{Proceedings 2000 ICRA. Millennium Conference. IEEE
  International Conference on Robotics and Automation. Symposia Proceedings
  (Cat. No.00CH37065)}, volume~2, 1207--1212 vol.2.
\newblock \doi{10.1109/ROBOT.2000.844763}.

\bibitem[{Bhattacharya and Agrawal(2000{\natexlab{b}})}]{897794}
Bhattacharya, S. and Agrawal, S.K. (2000{\natexlab{b}}).
\newblock Spherical rolling robot: a design and motion planning studies.
\newblock \emph{IEEE Transactions on Robotics and Automation}, 16(6), 835--839.
\newblock \doi{10.1109/70.897794}.

\bibitem[{Bloch(2003)}]{ambloch}
Bloch, A. (2003).
\newblock \emph{Nonholonomic Mechanics and Control}.
\newblock Springer-Verlag New York.

\bibitem[{Cai et~al.(2012)Cai, Zhan, and Xi}]{cai2012path}
Cai, Y., Zhan, Q., and Xi, X. (2012).
\newblock Path tracking control of a spherical mobile robot.
\newblock \emph{Mechanism and Machine Theory}, 51, 58--73.

\bibitem[{Crossley(2006)}]{crossley2006literature}
Crossley, V.A. (2006).
\newblock A literature review on the design of spherical rolling robots.
\newblock \emph{Pittsburgh, Pa}, 1--6.

\bibitem[{Dwaracherla et~al.(2019)Dwaracherla, Thakar, Vachhani, Gupta, Yadav,
  and Modi}]{8794742}
Dwaracherla, V., Thakar, S., Vachhani, L., Gupta, A., Yadav, A., and Modi, S.
  (2019).
\newblock Motion planning for point-to-point navigation of spherical robot
  using position feedback.
\newblock \emph{IEEE/ASME Transactions on Mechatronics}, 24(5), 2416--2426.
\newblock \doi{10.1109/TMECH.2019.2934789}.

\bibitem[{Hajos et~al.(2005)Hajos, Jones, Behar, and Dodd}]{hajos2005overview}
Hajos, G., Jones, J., Behar, A., and Dodd, M. (2005).
\newblock An overview of wind-driven rovers for planetary exploration.
\newblock In \emph{43rd AIAA aerospace sciences meeting and exhibit}, 244.

\bibitem[{Halme et~al.(1996)Halme, Schonberg, and Wang}]{509415}
Halme, A., Schonberg, T., and Wang, Y. (1996).
\newblock Motion control of a spherical mobile robot.
\newblock In \emph{Proceedings of 4th IEEE International Workshop on Advanced
  Motion Control - AMC '96 - MIE}, volume~1, 259--264 vol.1.
\newblock \doi{10.1109/AMC.1996.509415}.

\bibitem[{Kalita et~al.(2020)Kalita, Gholap, and
  Thangavelautham}]{kalita2020dynamics}
Kalita, H., Gholap, A.S., and Thangavelautham, J. (2020).
\newblock Dynamics and control of a hopping robot for extreme environment
  exploration on the moon and mars.
\newblock In \emph{2020 IEEE Aerospace Conference}, 1--12. IEEE.

\bibitem[{Kayacan et~al.(2012)Kayacan, Bayraktaroglu, and Saeys}]{decoupled}
Kayacan, E., Bayraktaroglu, Z.y., and Saeys, W. (2012).
\newblock Modeling and control of a spherical rolling robot: A decoupled
  dynamics approach.
\newblock \emph{Robotica}, 30(4), 671--680.
\newblock \doi{10.1017/S0263574711000956}.
\newblock \urlprefix\url{http://dx.doi.org/10.1017/S0263574711000956}.

\bibitem[{Lee and Park(2013)}]{lee2013design}
Lee, J. and Park, W. (2013).
\newblock Design and path planning for a spherical rolling robot.
\newblock In \emph{ASME International Mechanical Engineering Congress and
  Exposition}, volume 56246, V04AT04A028. American Society of Mechanical
  Engineers.

\bibitem[{Li and Liu(2011)}]{li2011design}
Li, T. and Liu, W. (2011).
\newblock Design and analysis of a wind-driven spherical robot with multiple
  shapes for environment exploration.
\newblock \emph{Journal of Aerospace Engineering}, 24(1), 135--139.

\bibitem[{Li et~al.(2012)Li, Wang, and Ji}]{li2012dynamic}
Li, T., Wang, Z., and Ji, Z. (2012).
\newblock Dynamic modeling and simulation of the internal-and external-driven
  spherical robot.
\newblock \emph{Journal of Aerospace Engineering}, 25(4), 636--640.

\bibitem[{Liu et~al.(2008)Liu, Sun, and Jia}]{liu2008family}
Liu, D., Sun, H., and Jia, Q. (2008).
\newblock A family of spherical mobile robot: Driving ahead motion control by
  feedback linearization.
\newblock In \emph{2008 2nd International Symposium on Systems and Control in
  Aerospace and Astronautics}, 1--6. IEEE.

\bibitem[{Mahboubi et~al.(2013)Mahboubi, Seyyed~Fakhrabadi, and
  Ghanbari}]{mahboubi2013design}
Mahboubi, S., Seyyed~Fakhrabadi, M.M., and Ghanbari, A. (2013).
\newblock Design and implementation of a novel spherical mobile robot.
\newblock \emph{Journal of Intelligent \& Robotic Systems}, 71(1), 43--64.

\bibitem[{Michaud and Caron(2002)}]{Michaud}
Michaud, F. and Caron, S. (2002).
\newblock Roball, the rolling robot.
\newblock \emph{Auton. Robots}, 12(2), 211--222.
\newblock \doi{10.1023/A:1014005728519}.
\newblock \urlprefix\url{http://dx.doi.org/10.1023/A:1014005728519}.

\bibitem[{Raura et~al.(2017)Raura, Warren, and
  Thangavelautham}]{raura2017spherical}
Raura, L., Warren, A., and Thangavelautham, J. (2017).
\newblock Spherical planetary robot for rugged terrain traversal.
\newblock In \emph{2017 IEEE Aerospace Conference}, 1--10. IEEE.

\bibitem[{{Rosen}(2000)}]{rosen}
{Rosen}, A. (2000).
\newblock {Modified Lagrange Method to Analyze Problems of Sliding and
  Rolling}.
\newblock \emph{Journal of Applied Mechanics}, 67, 697.
\newblock \doi{10.1115/1.1328088}.

\bibitem[{Tomik et~al.(2012)Tomik, Nudehi, Flynn, and
  Mukherjee}]{tomik2012design}
Tomik, F., Nudehi, S., Flynn, L.L., and Mukherjee, R. (2012).
\newblock Design, fabrication and control of spherobot: A spherical mobile
  robot.
\newblock \emph{Journal of Intelligent \& Robotic Systems}, 67(2), 117--131.

\bibitem[{Ylikorpi et~al.(2014{\natexlab{a}})Ylikorpi, Forsman, Halme, Saarinen
  et~al.}]{ylikorpi2014unified}
Ylikorpi, T., Forsman, P., Halme, A., Saarinen, J., et~al.
  (2014{\natexlab{a}}).
\newblock Unified representation of decoupled dynamic models for
  pendulum-driven ball-shaped robots.
\newblock In \emph{Proceedings of the 28th European Conference on Modelling and
  Simulation}. ECMS.

\bibitem[{Ylikorpi et~al.(2014{\natexlab{b}})Ylikorpi, Forsman, Halme
  et~al.}]{ylikorpi2014gyroscopic}
Ylikorpi, T., Forsman, P., Halme, A., et~al. (2014{\natexlab{b}}).
\newblock Gyroscopic precession in motion modelling of ball-shaped robots.
\newblock In \emph{Proceedings of the 28th European Conference on Modelling and
  Simulation}. ECMS.

\bibitem[{Yoon et~al.(2011)Yoon, Ahn, and Lee}]{yoon2011spherical}
Yoon, J.C., Ahn, S.S., and Lee, Y.J. (2011).
\newblock Spherical robot with new type of two-pendulum driving mechanism.
\newblock In \emph{2011 15th IEEE International Conference on Intelligent
  Engineering Systems}, 275--279. IEEE.

\end{thebibliography}

                                                   







\end{document}